\documentclass{article} 
\usepackage{iclr2026_conference_preprint,times}

\usepackage{amsmath,amsfonts,bm}

\def\eqref#1{equation~\ref{#1}}

\def\1{\bm{1}}

\DeclareMathAlphabet{\mathsfit}{\encodingdefault}{\sfdefault}{m}{sl}
\SetMathAlphabet{\mathsfit}{bold}{\encodingdefault}{\sfdefault}{bx}{n}

\usepackage{hyperref}
\usepackage{url}
\usepackage{float}
\usepackage{booktabs}
\usepackage{tabularx}
\usepackage{graphicx}
\usepackage{xspace}

\newcommand{\bfX}{\mathbf{X}}
\newcommand{\bfx}{\mathbf{x}}
\newcommand{\mcD}{\mathcal{D}}
\newcommand{\stable}{\emph{stable}\xspace}
\newcommand{\standard}{\emph{standard}\xspace}
\newcommand{\ensemble}{\emph{ensemble}\xspace}

\title{Bootstrapping-based Regularisation for Reducing Individual Prediction Instability in Clinical Risk Prediction Models}

\iclrfinalcopy

\author{
Sara Matijevic \\
Nuffield Department of Women's and \\
Reproductive Health, \\
University of Oxford, \\
Oxford, UK \\
\And
Christopher Yau\thanks{Correspondence to: christopher.yau@wrh.ox.ac.uk} \\
Nuffield Department of Women's and \\
Reproductive Health, \\
University of Oxford \\
Oxford, UK \\
Health Data Research UK, \\
Gibbs Building, 215 Euston Road \\
London, UK
}

\begin{document}

\maketitle

\begin{abstract}
Clinical prediction models are increasingly used to support patient care, yet many deep learning–based approaches remain unstable, as their predictions can vary substantially when trained on different samples from the same population. Such instability undermines reliability and limits clinical adoption. In this study, we propose a novel bootstrapping-based regularisation framework that embeds the bootstrapping process directly into the training of deep neural networks. This approach constrains prediction variability across resampled datasets, producing a single model with inherent stability properties. We evaluated models constructed using the proposed regularisation approach against conventional and ensemble models using simulated data and three clinical datasets: GUSTO-I, Framingham, and SUPPORT. Across all datasets, our model exhibited improved prediction stability, with lower mean absolute differences (e.g., 0.019 vs. 0.059 in GUSTO-I; 0.057 vs. 0.088 in Framingham) and markedly fewer significantly deviating predictions. Importantly, discriminative performance and feature importance consistency were maintained, with high SHAP correlations between models (e.g., 0.894 for GUSTO-I; 0.965 for Framingham). While ensemble models achieved greater stability, we show that this came at the expense of interpretability as each constituent model used predictors in different way. By regularising predictions to align with bootstrapped distributions, our approach allows prediction models to be developed that achieve greater robustness and reproducibility without sacrificing interpretability. This method provides a practical route toward more reliable and clinically trustworthy deep learning models, particularly valuable in data-limited healthcare settings.
\end{abstract}

\section{Introduction}

Clinical prediction models (CPMs) quantify the risk of individual patient outcomes and assess the likelihood of diagnostic or prognostic events. These models can aid clinicians in making informed decisions about patient care by systematically incorporating multiple predictive factors including demographic variables (age, weight, family history, comorbidities), biological measurements (blood pressure, biomarkers), and diagnostic test results. Established examples such as QRisk \citep{hippisley2017development} for cardiovascular risk assessment and EuroSCORE \citep{nashef2012euroscore} for cardiac surgery risk evaluation demonstrate the practical utility of these approaches in clinical practice.

Prediction models can be constructed using statistical methods such as logistic (binary outcomes) or Cox regression (time to event outcomes). More recently, machine learning-based approaches, such as random forests and deep neural networks, have been adopted to allow more flexible relationships between predictors and outcomes. 

Recently, the topic of \emph{prediction model instability} has gained renewed interest  \citep{riley2023clinical}. These works have highlighted that the predictive properties of a  prediction model are tied to the sample and size of the data used to develop it. If a different sample of the same size were drawn from the same overarching population, the resulting prediction model could differ substantially, even when the same development methods and architectures are applied. These notions are also well-known in machine learning for models trained according to \emph{empirical risk minimisation} (ERM) principles where, in the absence of knowledge of the true data distribution, it is necessary to adopt an approximation based on the empirical sample itself. 

\cite{riley2023clinical} recommends, particularly when development data sizes are small, that the individual-level prediction variability be reported by bootstrapping the development data and creating multiple versions of the prediction model. The machine learning community has addressed the same problem through a related approach which are termed \emph{ensemble methods}. Utilising similar bootstrapping ideas but instead of merely reporting variability, \emph{bagging} methods \citep{bagging1} aggregate the outputs of multiple models to produce a single, more stable prediction. This is based on the principle that ``different models will usually not make all the same errors on the test set" \citep{goodfellow2016deep}.

Interpretability has also become a critical concern for deep learning prediction models, which are often viewed as `black boxes' due to the difficulty of understanding the encoded predictor–outcome relationships \cite{stiglic2020interpretability}. Feature importance measures such as Shapley values provide one way to address this. However, when ensemble methods are used, the feature importance attributions can differ across models within the ensemble, making interpretation more difficult since each model may utilise the predictive features differently. Since clinical practice requires both reliable predictions and consistent explanations, maintaining interpretability alongside stability could be seen to be a more coherent approach.

In this work, we propose a novel prediction model training criterion which embeds a bootstrapping process directly within the training of deep neural networks. This approach enables us to regularise a model to achieve the stability benefits typically associated with ensembles, while maintaining the interpretability advantage of having only a single model. We demonstrate how to construct an efficient, generalisable framework that is applicable to any differentiable binary outcome prediction model and demonstrate using simulated and real data examples.

\section{Methodology}
\subsection{Background}

We first provide a mathematical specification of the stability problem. For this work, we will focus on binary prediction problems and leave generalisations to more complex scenarios, e.g. time-to-event, as further work.

Let $Y \in \{0, 1\}$ denote a binary random variable representing prediction outcome. We suppose that the probability of an event, given a vector of $P$ predictors $\bfX \in \mathbb{R}^P$, is given  $P(Y=1|\bfX=\bfx) = f_\theta(\bfx)$ where $f$ is some unknown function with model parameters $\theta$. A training or development data set is available, $\mcD = \{ \{ \bfx_1, y_1 \}, \dots, \{ \bfx_N, y_N \} \}$, where the observations are independent and identically distributed and drawn from a population $\mathcal{P}$. Using the data, the goal is to identify an estimate $\hat{\theta}$ for the model parameters in order to produce a predictive function $P(Y|\bfX = \bfx^*) = f_{\hat{\theta}}(\bfX = \bfx^*)$ given a new set of predictors $\bfx^*$.

In this work, we will adopt a maximum likelihood-based learning approach so that $\hat{\theta} = \arg \max_{\theta} \ln P(\mcD|\theta)$ which can be equivalently expressed as the minimisation of the following loss function:
\begin{align}
	\label{eq:bce_loss}
	L_{\theta}(\mcD) 
		&= -\ln P(\mcD|\theta) = -\sum_{i=1}^N \Big[ y_i \log f_\theta(\bfx_i)  + (1 - y_i) \log (1 - f_\theta(\bfx_i)) \Big ].
\end{align}
This also coincides with empirical risk minimisation (ERM) using a binary cross-entropy loss.

\begin{figure}[H]
	\centering
	\includegraphics[width=0.55\textwidth]{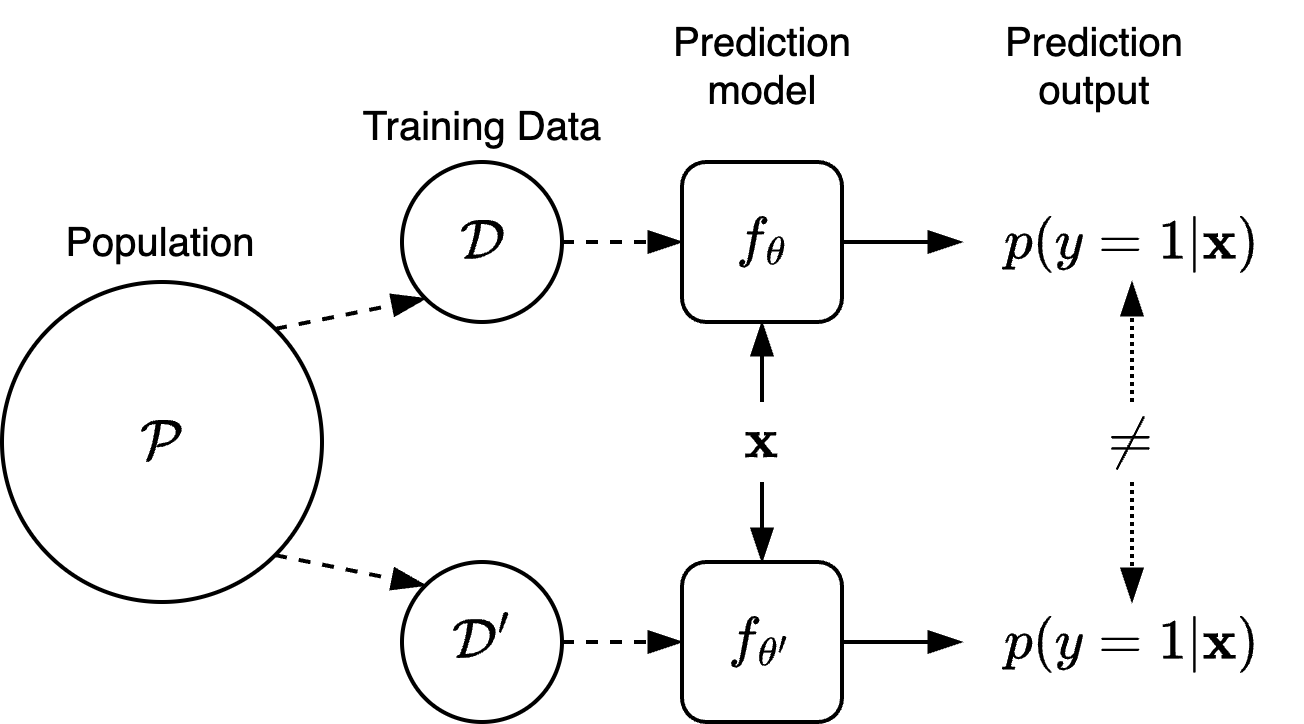}
	\caption{{\bf Approach Overview.} Prediction models trained on different development data sets $\mcD/\mcD'$, even drawn from the same population $\mathcal{P}$, can lead to models ($f_\theta/f_{\theta'}$) which produce different risk probabilities at the individual level.}
	\label{fig:schematic}
\end{figure}

As the estimator $\hat{\theta}$ is dependent on $\mcD$, it is a random quantity. Consequently, suppose a second identically-sized training sample could be drawn $\mcD' \sim \mathcal{P}$ and a prediction model trained on this to obtain $f_{\theta'}(\bfx)$. In general, the predictions from the resultant models given the same input will not be the same, $f_\theta(\bfX^*) \neq f_{\theta'}(\bfX^*)$, given finite samples $\mcD$ and $\mcD'$ (Figure \ref{fig:schematic}). In this work, individual-level prediction (ILP) instability refers to the difference in prediction output between a model derived from the training data and an alternative model that could have been derived from another similarly sized data set drawn from the same population. Note that in ensemble learning such differences are also referred to by the term \emph{diversity} highlighting the diverse models that could be learnt from different training sets drawn from the same population.

\cite{riley2023clinical} recommends reporting the variance associated with ILPs by bootstrapping the training data, fitting a model to each bootstrapped sample and computing the variance in ILPs across bootstrapped models. In contrast, in ensemble learning, a reduction in the variance of the parameter estimates and the ILPs can be achieved by aggregating across bootstrapped model predictions. Both approaches can be considered to be post-hoc approaches which retrospectively recognise (and correct for) the model diversity issue. 

\subsection{Stability-based regularisation}
Rather than reporting or correcting for instability post hoc, we introduce our novel stability-based regularisation approach which actively incorporates model diversity during training leading to the production of a \emph{single} prediction model. We denote $\mcD^b = \{ \bfX^b, Y^b \} \sim B(\mcD)$ as a bootstrap sample of the original training data where we use $B(\cdot)$ to denote a bootstrap sampling operation. Next let $\hat{\theta}_b$ be the maximum likelihood estimates of the model parameters that would be derived by training with a bootstrapped sample $\mcD^b$. This allows us to define the absolute difference between the (log) prediction probabilities given by the \emph{target} model  $f_\theta(\bfx)$, which is the model we are training and based on the original training data $\mathcal{D}$, and an alternative model which could have been developed based on a bootstrapped dataset $\mathcal{D}^b$ as
\begin{equation}
	\label{eq:diff_metric}
	d(f_{\hat{\theta}^b}(\bfx), f_\theta(\bfx)) = || \log f_{\hat{\theta}^b}(\bfx) - \log f_\theta(\bfx) || .
\end{equation}

We now propose a new penalised likelihood function:
\begin{align}
	\label{eq:regularised_bce_loss}
	\mathcal{R}(\theta) = L_{\theta}(\mcD) + \frac{\lambda}{N} \sum_{i=1}^N \mathbb{E}_{B(\mcD)} [ d(f_{\hat{\theta}^b}(\bfx_i), f_\theta(\bfx_i)) ]
\end{align}
where $\lambda$ is a regularisation penalty term and the expectation in the latter component with respect to the bootstrapping distribution. The regularisation term measures the expected difference in predictions by the target model $f_\theta(\bfx)$ versus the other possible models that could have been generated under different (bootstrapped) variations of the training data. Optimising to find $\hat{\theta} = \arg \max_{\theta} \mathcal{R} (\theta)$ therefore balances fit to data imposed by $L_{\theta}(\mcD)$ while accounting for diversity through the regulariser on ILPs across bootstrapped models.

In implementation, we approximate the expectation by aggregating over $M$ bootstrapped models which are pre-computed:
\begin{equation}
	\mathbb{E}_{B(\mcD)} [ d(f_{\hat{\theta}^b}(\bfx_i), f_\theta(\bfx_i)) ] \approx \frac{1}{M} \sum_{m=1}^M d(f_{\hat{\theta}^{b_m}}(\bfx_i), f_\theta(\bfx_i))
\end{equation}
This reduces the computational burden so that the $f_{\hat{\theta}^{b_m}}$'s do not need to be trained nor evaluated on-the-fly during the training of the target model $f_\theta$. Model training for the target model, when this is defined by a deep neural network, can then follow conventional gradient-based optimisation.

Table \ref{tab:stability_comparison} summarises the different approaches.

\begin{table}[H]
\centering
\caption{{\bf Methods Comparison.} Comparison of approaches for addressing prediction instability in clinical risk models.}
\label{tab:stability_comparison}

\begin{tabularx}{\textwidth}{p{3.2cm} X X}
\toprule
\textbf{Approach} & \textbf{Mechanism} & \textbf{Strengths and Limitations} \\
\midrule
\textbf{Statistical bootstrapping} (Riley et al., 2023)
& Repeatedly resample training data to \textit{report} individual-level prediction variability.
& Provides insight into instability but does not change the trained model. \\
\midrule
\textbf{Bagging / Ensembles} (Breiman, 1996)
& Train multiple models on bootstrap samples and average predictions.
& Improves stability but reduces interpretability, since feature contributions differ across models. \\
\midrule
\textbf{Conventional regularisation} (e.g., L1/L2, dropout)
& Penalise weights or activations to reduce variance and overfitting.
& Improves generalisation but does not explicitly address prediction instability across datasets. \\
\midrule
\textbf{Our stability-based regularisation}
& Embed a bootstrapping process directly into the training objective. Penalise deviations between predictions from the model and those that would arise from bootstrapped resamples.
& Produces a \textit{single model} with ensemble-like stability while retaining interpretability. \\
\bottomrule
\end{tabularx}
\end{table}

\subsection{Experimental Setup}
To demonstrate the approach, in the following experiments, we use a deep neural network (DNN) architecture for $f$. The network consists of an input layer corresponding to the number of predictor variables, followed by two hidden layers with 64 and 32 neurons, respectively, using ReLU activation functions. The output layer consists of a single neuron with a sigmoid activation function to produce a probability estimate in the range $(0,1)$. The model is trained using stochastic gradient descent with an Adam optimiser to optimise the parameters $\theta$. In each iteration, we use 100 bootstrapped models to approximate the expectations, which is the minimum number recommended by \citet{stability}, but these are randomly sampled from a larger pool of a total of 200 bootstrapped samples for which the corresponding prediction models were pre-computed. 

In each experiment, we refer to the model constructed using our regularisation approach as the \stable model. The \emph{standard} prediction model refers to the conventional prediction model fit once on the training data in the usual way. The third model named \ensemble refers to a bagged model aggregated using 200 bootstrapped samples. As a consequence, the computational requirements of the \ensemble and \stable models are almost exactly the same since both require computing 200 prediction models on 200 bootstrapped data sets. Our approach requires one training cycle for the final \stable model.

\subsection{Datasets}

\paragraph{Simulation study}We first created a simulated dataset to test the two models. The dataset consisted of $n=4,000$ samples, each characterized by $p=15$ predictor variables, including binary, continuous, and noise features. Two binary features were generated from a Bernoulli distribution with equal probability, such that $X_{\text{bin}} \sim \text{Bernoulli}(0.5)$. Ten continuous features carrying meaningful predictive information were sampled from a standard normal distribution, $X_{\text{imp}} \sim \mathcal{N}(0,1)$. Finally, three noise features were included to introduce variability, drawn from a uniform distribution, $X_{\text{noise}} \sim \text{Uniform}(-1,1)$. The final predictor matrix, $X_{\text{sim}}$, was obtained by concatenating these three groups of features.

The probability of an event occurring was then computed using the logistic function:
\begin{equation*}
P(Y_{\text{sim}} = 1 \mid X_{\text{imp}}) = \frac{1}{1 + \exp(-X_{\text{imp}} \beta)}.
\end{equation*}
where \(\beta \sim \text{Uniform}(3.0, 6.0) \).

\paragraph{GUSTO-I dataset}The GUSTO-I dataset was collected as part of the Global Utilization of Streptokinase and Tissue Plasminogen Activator for Occluded Coronary Arteries (GUSTO-I) trial, a large-scale randomised controlled trial conducted in the early 1990s with over 41,000 acute myocardial infarction (AMI) patients \citep{gusto}. The final dataset includes 40,830 participants with 2,851 deaths by 30 days, which was defined as the outcome variable. The parameters used in the analysis below were those previously found to be statistically significant in other research.  These included predictors such as age (years), sex (0 = male, 1 = female), systolic blood pressure, weight, pulse rate, Killip classification (I-IV), history of hypertension (0 = no, 1 = yes), history of myocardial infarction (0 = no, 1 = yes), anterior infarction (0 = no, 1 = yes), infarction localization (anterior, inferior, other), smoking status (current, never, former), and ST elevation on ECG (number of leads).

\paragraph{Framingham Dataset}The Framingham dataset originates from the Framingham Heart Study, a longitudinal epidemiological study investigating cardiovascular disease (CVD) risk factors \citep{fram}. The dataset used in this analysis is derived from a later phase of the study and focuses on predicting coronary heart disease (CHD) outcomes based on established cardiovascular risk factors. The final dataset consists of 4,434 participants who were followed over time, with the occurrence of CHD recorded as a binary outcome variable. The variables used in the analysis were: sex (0 = male, 1 = female), age (years), systolic blood pressure (SBP, mmHg), diastolic blood pressure (DBP, mmHg), serum cholesterol levels (mg/dL) and body mass index (BMI, kg/m\(^2\)).

\paragraph{SUPPORT Dataset}SUPPORT (Study to Understand Prognoses and Preferences for Outcomes and Risks of Treatments) dataset was collected during a multi-center study aimed at understanding outcomes and decision-making for seriously ill hospitalised patients. The dataset contains 9,103 individuals that were enrolled in the study from five academic centers in the U.S. and patient outcomes include various diseases such as acute respiratory failure, congestive heart failure, and cancers. The dataset comprises 17 variables that cover diagnosis, age, sex, physiologic measures, and other comorbidities.

\subsection{Metrics and Plots}
 
\paragraph{MAD} To evaluate the stability of model predictions, we computed the mean absolute difference (MAD) between the test set predictions of the \standard, \stable  and \ensemble prediction models and the median predictions of the bootstrapped models. This metric quantifies how much individual predictions deviate from the central tendency of the bootstrapped distribution.

\paragraph{Prediction deviation} We also calculated, for each participant, the proportion of cases in which the \stable or \ensemble model’s prediction lay closer to the bootstrapped median than the \standard model’s prediction. To assess statistical significance, we used the empirical bootstrap distribution to compute a p‐value for each participant. More specifically, we measured the probability of observing an absolute deviation from the bootstrapped median at least as large as the one produced by the standard model.

\paragraph{AUC} To assess the discriminative ability of the models, we computed the area under the receiver operating characteristic (ROC) curve (AUC) for all three models on the held-out test set. The AUC measures how well a model differentiates between outcome classes, with a higher value indicating better discrimination.

\paragraph{SHAP} Furthermore, in order to analyse if the focus on prediction stability has an impact on feature selection, we used SHapley Additive exPlanations (SHAP) to quantify the contribution of each predictor to the model’s output \citep{roth1988introduction}. We computed the Spearman correlation to measure the overall agreement in feature‐importance rankings assigned by the two models across all predictions. Additionally, we evaluated per‐participant SHAP correlation, which quantified how similarly the two models attribute importance to features for each individual, as well as per‐feature SHAP correlation, which measured the consistency of feature importance across participants. Lastly, we also computed the SHAP values on for each of the twenty bagged models that comprise the \ensemble model and then plotted per‐feature violin plots of their SHAP standard deviations, thereby visualizing how variable each feature’s importance is across the ensemble members.

\section{Results}
\paragraph{Simulation Data} 

\begin{figure}[!ht]
    \centering
    \includegraphics[width=0.77\textwidth]{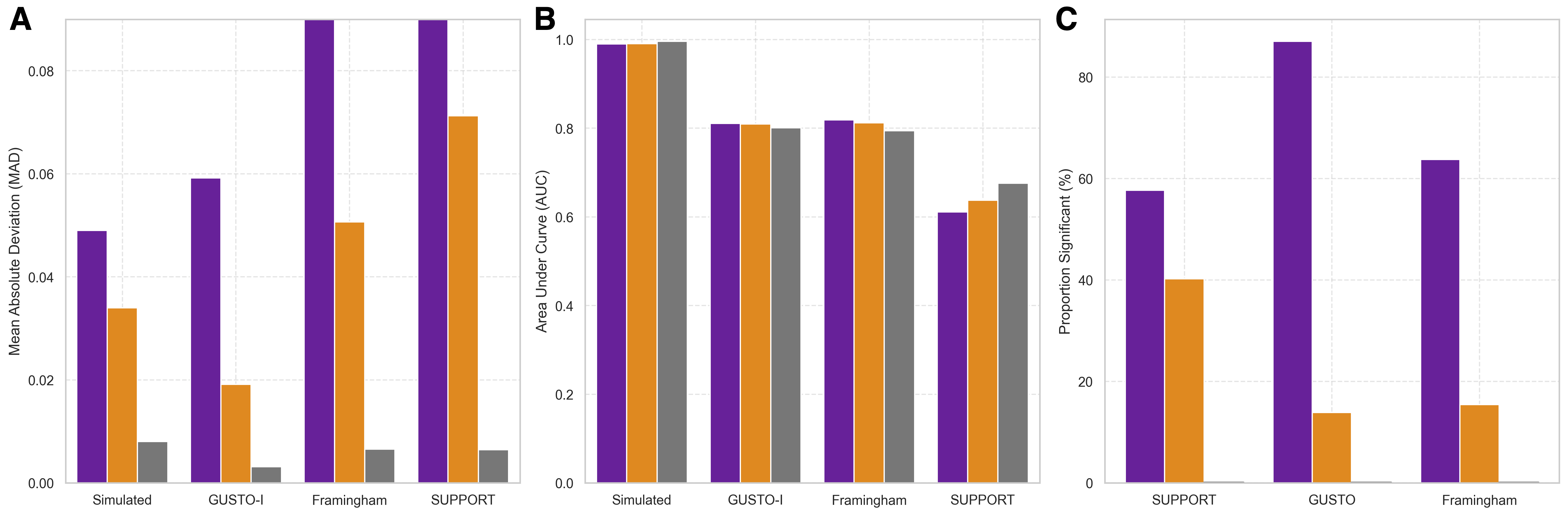}
    \caption{{\bf Prediction model stability}. Comparison of {\bf (A)} prediction stability (MAD), {\bf (B)} predictive performance (AUC), and {\bf (C)} the proportion of significantly deviating predictions for the {\color{violet}\standard} (violet), {\color{orange}\stable} (orange), and {\color{gray}\ensemble} (gray) models across the simulated, GUSTO‐I, Framingham, and SUPPORT datasets.}
    \label{fig:prediction_model_stability}
\end{figure}

\begin{figure}[h]
    \centering
    \includegraphics[width=0.8\textwidth]{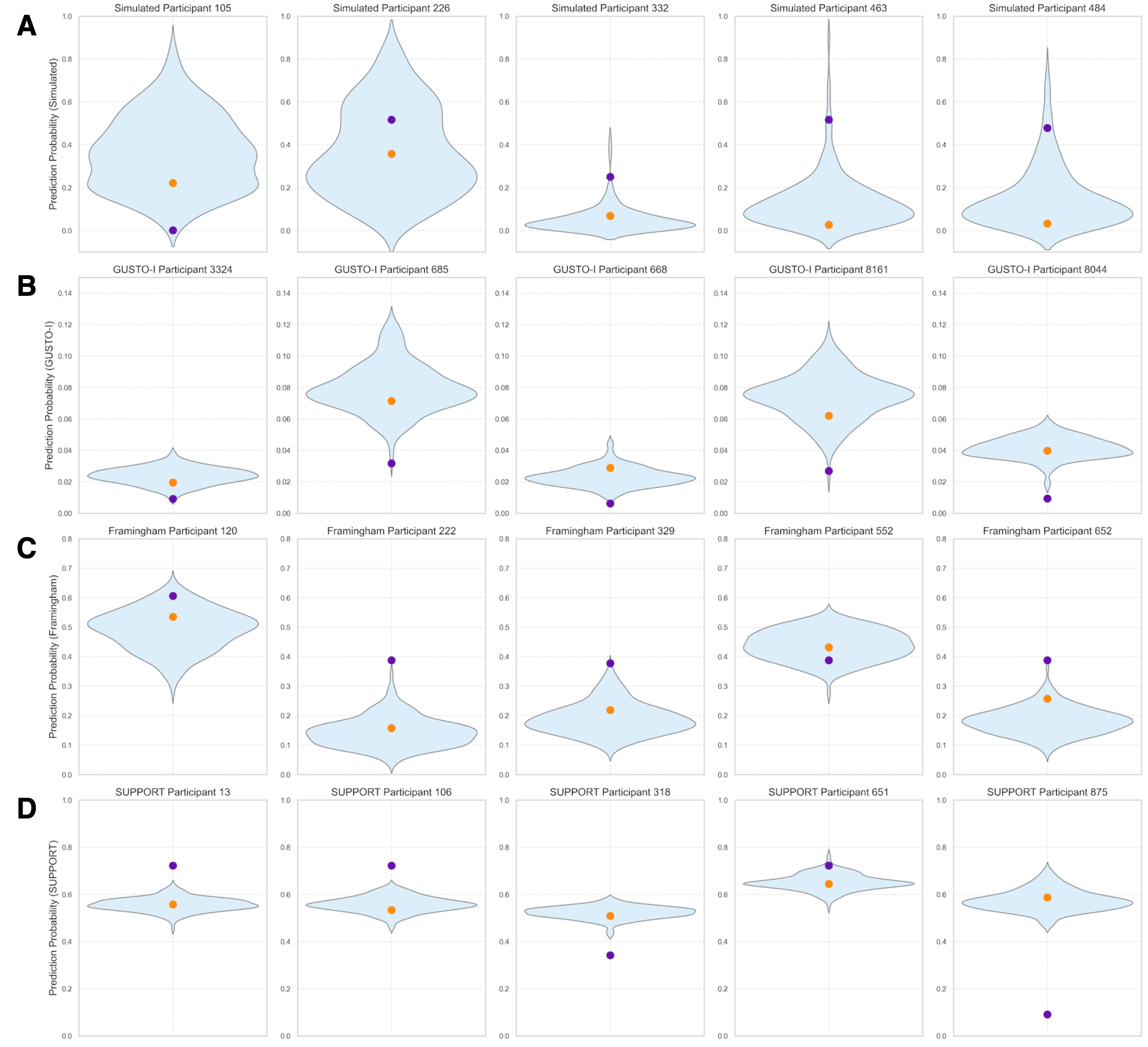}
    \caption{{\bf Individual prediction stability}. Individual-level predictions for selected participants from {\bf (A)} the simulated, {\bf (B)} GUSTO-I, {\bf (C)} Framingham and {\bf(D)} SUPPORT datasets. The violin plots display the distribution of predictions from 200 bootstrapped models while the dots indicate the {\color{violet}\standard}  (violet) and {\color{orange}\stable} (orange)predictions.}
    \label{fig:individual_level_predictions}
\end{figure}

\begin{figure}[h]
    \centering
    \includegraphics[width=0.8\textwidth]{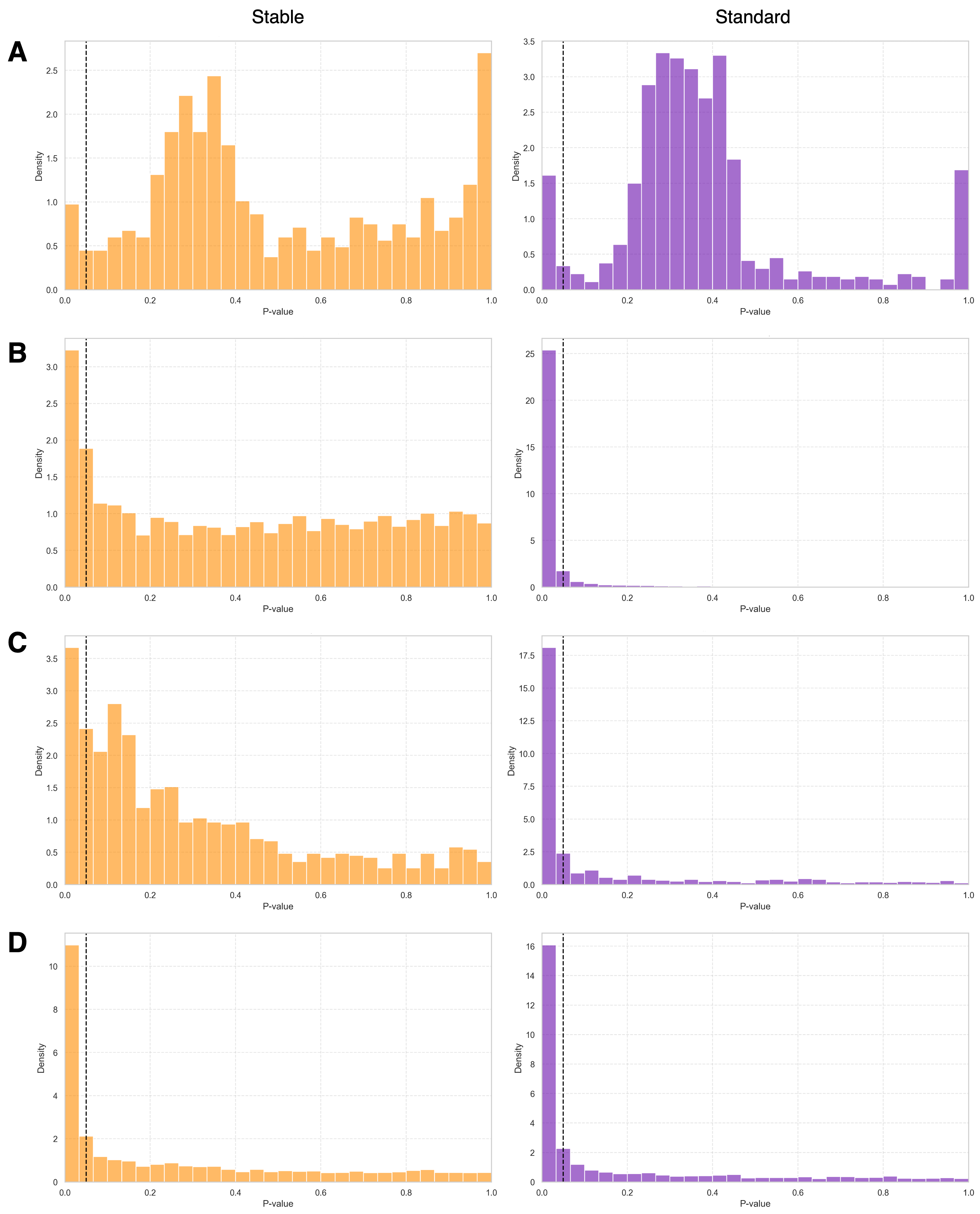}
    \caption{{\bf Prediction deviation}. Histograms of p-values for the significance of individual-level prediction deviations from the bootstrapped median for {\color{orange}\stable} (orange) and {\color{violet}\standard} (violet) models for the {\bf (A)} simulated, {\bf (B)} GUSTO-I, {\bf (C)} Framingham and {\bf (D)} SUPPORT datasets. The dashed line shows a 0.05 significance threshold.}
    \label{fig:pvals}
\end{figure}

We first considered the analysis of the simulated dataset. Figure \ref{fig:prediction_model_stability}A shows the relative performance of the three approaches. The \standard model had a MAD of 0.048 compared to 0.034 for the \stable model demonstrating the effect of the regularisation on reducing variability in the ILPs. The \ensemble model had the lowest MAD but this is to be expected since the aggregation in the bagging approach acts to directly optimise for MAD. All three models achieved similar AUC scores indicating that the regularisation had no direct impact on predictive performance (Figure \ref{fig:prediction_model_stability}B). Figure \ref{fig:individual_level_predictions}A illustrates individual examples where the \stable model's predictions better align with the peaks of the bootstrapped prediction distributions. Furthermore, only 3.88\% of \stable model's predictions were classified as significantly deviating from the median of the bootstrapped distribution compared to 6.12\% for the \standard model. This can be seen in the p-value distribution plots (Figure \ref{fig:pvals}A) which show the proportion of predictions falling below the 0.05 significance threshold.

\paragraph{Real Data} 

We next considered the real world data examples. On the GUSTO-I dataset, the \stable model achieved a MAD of 0.019 versus 0.059 for the \standard model (Figure \ref{fig:prediction_model_stability}A). While the smaller Framingham dataset, the \stable model had a MAD of 0.057 compared to 0.088. On the SUPPORT dataset, the \stable model had a MAD of 0.0712 compared to 0.0923 for the \standard model. There was no substantive difference in the AUC achieved between approaches (Figure \ref{fig:prediction_model_stability}B) again demonstrating that the regularisation did not impact on predictive performance. Across all three data sets, the proportion of individual-level predictions from the \standard model that were significantly different from the median of the bootstrapped distribution was much greater than for the \stable model (Figure \ref{fig:prediction_model_stability}C). This is further illustrated in the individual prediction plots in Figure \ref{fig:individual_level_predictions}B--D where the \stable model predictions are shown to be consistently closer to the median of the bootstrapped distributions than those of the \standard model. 

More precisely, for the GUSTO-I dataset, 13.89\% of the \stable model's predictions were classified as significantly deviating from the median of the bootstrapped distribution compared to 87.06\% for the \standard model. Similarly, for the Framingham dataset, 21.35\% of \stable model's predictions were identified as significant, compared to 55.04\% for the \standard model. For the SUPPORT dataset, 40.21\% of the \stable model’s predictions were classified as significantly deviating from the bootstrap median, compared to 57.71\% for the \standard model. This is further characterised in Figure \ref{fig:pvals}, where the proportion of predictions falling below the significance threshold for each dataset is shown. Here, the \standard model produces more ILPs which are significantly different from the bootstrapped median than the \stable model. Additionally, the global Spearman correlation, which was computed over all SHAP values flattened across participants and features, was 0.862 for GUSTO-I and 0.923 for Framingham, indicating very strong overall concordance in feature attributions. For the SUPPORT dataset, this correlation was more moderate at 0.529, pointing to a weaker alignment between the two models also reflected in the lower AUC for the SUPPORT dataset.

\paragraph{Ensemble models} In all instances, the \ensemble model showed strong predictive performance (AUC) as well as the greatest stability (MAD). So what is the purpose of our approach? The bagging models has the lowest MAD of all three models and almost no significantly deviating predictions from the median of the bootstrapped distribution. This is to be expected since the bagging operation performs mean aggregation across the bootstrapped models to give each ILP. However, bagging achieves this maximal ILP stability at the expense of model interpretation which is demonstrated in Appendix A, which shows the variation in SHAP values across ensemble model constituents or each of the real data sets.

\begin{figure}[h]
    \centering
    \includegraphics[width=\textwidth]{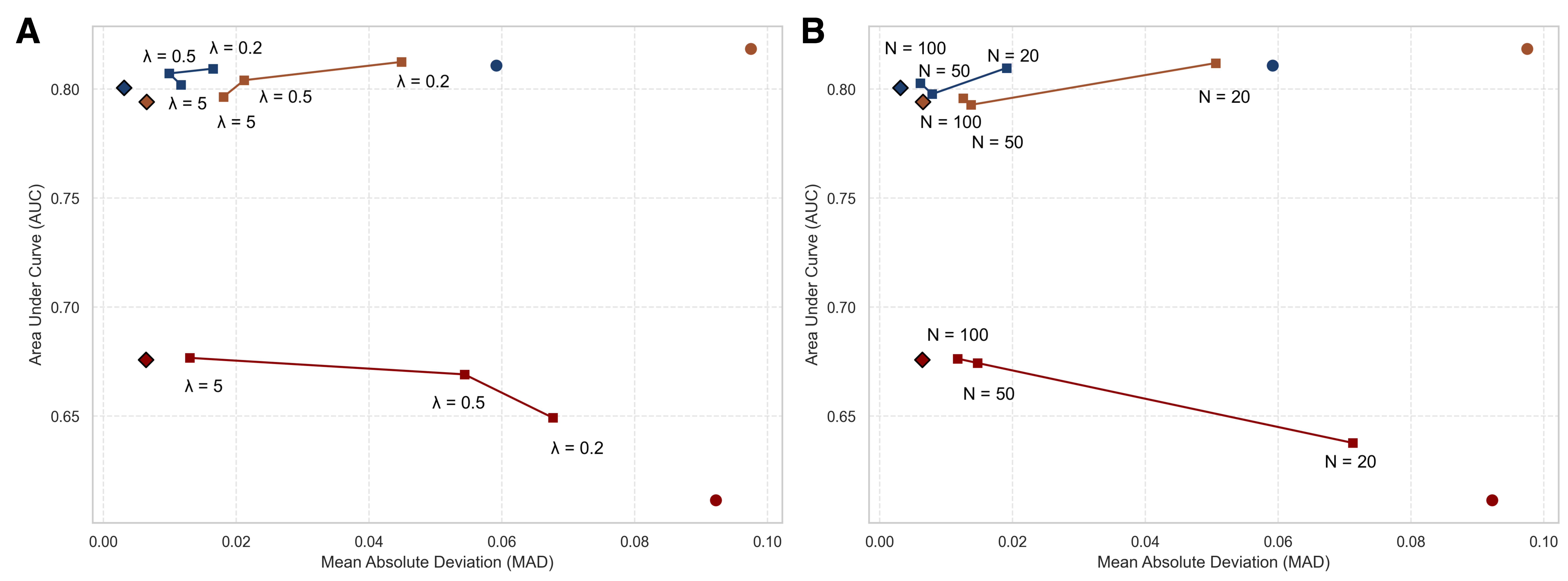}
    \caption{{\bf Hyperparameter Sensitivity}. Impact on AUC and MAD from changes in {\bf (A)} $\lambda$ and {\bf (B)} number of bootstrapped models in the \stable model  for {\color{blue}GUSTO}, {\color{brown}Framingham} and {\color{red}SUPPORT}. The diamond and the dot represent the \ensemble and \standard model respectively.}
    \label{fig:hyperparameter}
\end{figure}

\paragraph{Hyperparameter sensitivity} We next explored how the \stable model’s behaviour evolves when we adjust key hyperparameters: $\lambda$ and the number of bootstrapped samples $M$. As seen in Figure \ref{fig:hyperparameter}A, when the regularisation strength $\lambda$ increases, there is a further reduction in MAD tending toward the \ensemble model value. Since $\lambda$ determines the relative weight of the regularisation versus the data fitting term in the loss function, a large value weights heavily in favour of minimising ILP differences across models. Therefore our \stable model converges toward making ILPs which are close to the \ensemble model though it will never be exactly the same since, as we have shown, the \ensemble approach allows aggregation across models which use predictors differently while our \stable model is constrained to one predictor configuration. We finally investigated the effect of the number of bootstrapped samples used for simulation. Figure \ref{fig:hyperparameter}B shows the use of 20, 50 and 100 samples with $\lambda = 0.1$. As expected, while a larger number of bootstraps improved stability, the use of 20 samples was already sufficient to improve stability over a standard prediction model with no regularisation. Methods for dataset-specific optimisation of both parameters maybe desirable and we leave this for future work.

\section{Discussion}
This study presents a novel bootstrapping-based regularisation framework for improving the individual-level prediction stability of clinical risk models. By integrating a bootstrapping process directly into the model’s training objective, the approach explicitly penalises deviations between predictions generated from the original dataset and those expected from bootstrapped resamples. Across simulated and empirical datasets (GUSTO-I, Framingham, and SUPPORT), this stability-based regularisation consistently reduced prediction variability, as shown by lower mean absolute differences (MAD), while maintaining equivalent discriminative performance (AUC). The results suggest that stability can be substantially enhanced without loss of predictive accuracy or interpretability.

Compared with conventional regularisation methods such as L1/L2 penalties, or dropout \cite{srivastava2014dropout}, which aim to control overfitting at the parameter level, the proposed method operates directly at the prediction level. This distinction allows the model to learn representations that are inherently robust to sampling variation rather than simply constrained in complexity. Similarly, ensemble or bagging approaches achieve stability by averaging predictions from multiple models, but they do so at the expense of interpretability, since feature attributions vary across ensemble constituents. In contrast, using the proposed regularisation achieves ensemble-like robustness within a single interpretable model - a crucial property for clinical implementation, where both consistent predictions and transparent reasoning are required.

Our concept of prediction stability is also closely related to both deep ensembles \cite{lakshminarayanan2017simple} and Bayesian neural networks \cite{lampinen2001bayesian}. Deep ensembles achieve robustness by averaging predictions across multiple neural networks trained with different data resamples or initialisations, thereby smoothing over the variability inherent in model training. Our method can be viewed as a single-model analogue of this principle, embedding an implicit ensemble effect through a bootstrapping-based regularisation term. Similarly, Bayesian neural networks promote stability by integrating predictions over the posterior distribution of parameters, thereby quantifying epistemic uncertainty. While our approach does not perform probabilistic inference, it represents a frequentist counterpart that penalises divergence from the empirical bootstrap distribution. Future work could explore hybrid frameworks combining stability-based regularisation with Bayesian or ensemble uncertainty estimation to achieve both stable and well-calibrated clinical predictions. 

A challenge with all these approaches is that improvements in stability attract greater computational demands. In our experiments, our computational costs were comparable to ensemble methods as we could pre-compute bootstrapped models. Further optimisation may be needed for high-dimensional data where models for handling such data modalities are more substantive. The sensitivity analysis highlights that appropriate tuning of the regularisation strength ($\lambda$) and the number of bootstrap samples ($M$) is essential to balance stability and flexibility. Extensions to non-tabular data types, such as medical imaging or longitudinal health records, and integration with uncertainty quantification approaches (e.g., Bayesian or conformal prediction) would be areas of interesting further investigation.

In summary, this bootstrapping-based regularisation approach offers a practical and interpretable pathway to enhance prediction stability in clinical models, complementing existing methods for improving model robustness and reproducibility in medical research.

\newpage

\bibliography{iclr2026_conference}
\bibliographystyle{iclr2026_conference}

\newpage
\appendix
\section{Appendix}

\begin{figure}[h]
    \centering
    \includegraphics[width=0.8\textwidth]{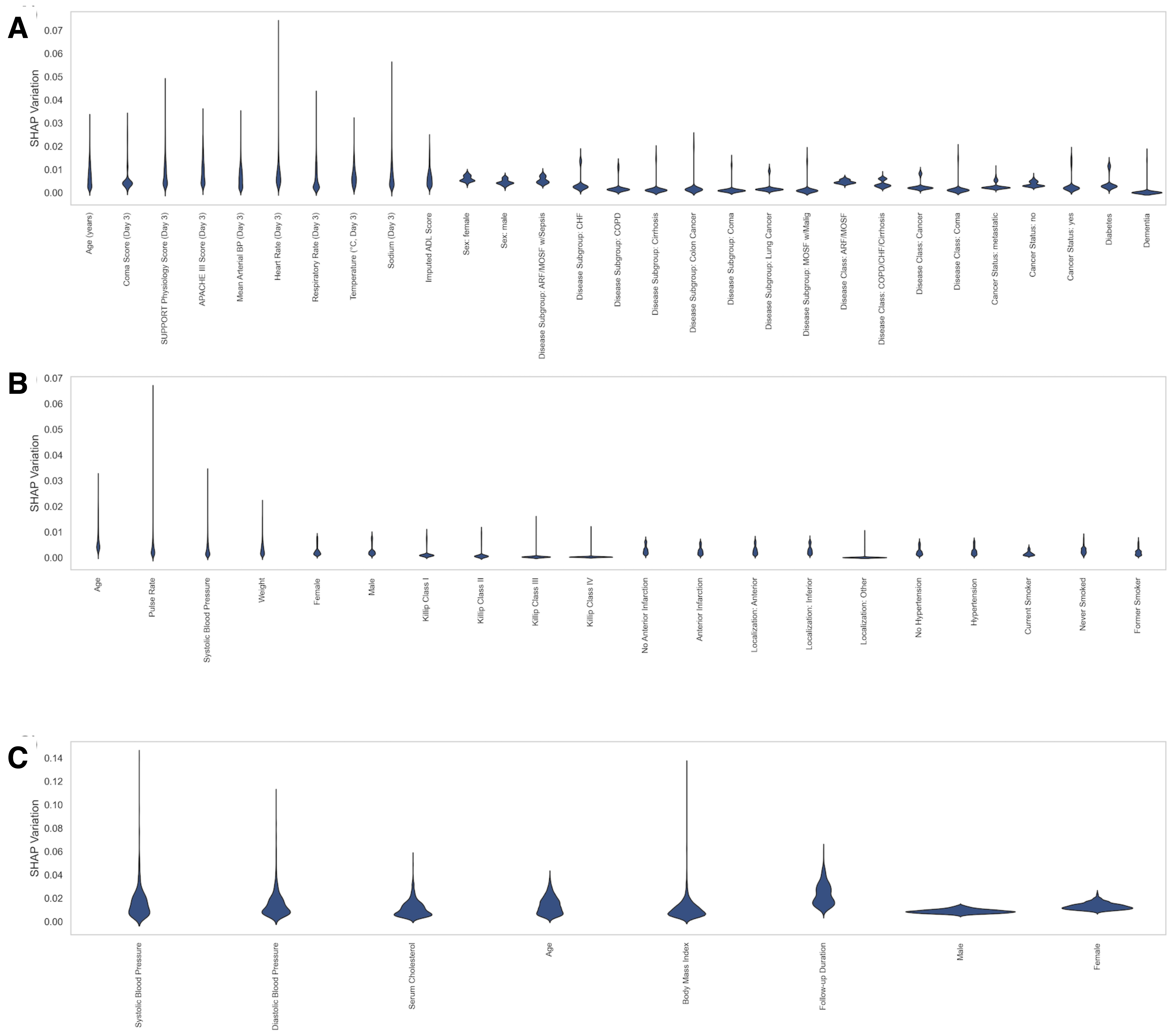}
    \caption{{\bf Feature-attribution score distribution across ensemble model constituents.}.Violin plots showing SHAP value variation for each feature in {\bf (A)} SUPPORT, {\bf (B)} GUSTO-I, and {\bf (C)} Framingham datasets.}
    \label{fig:ensembleshap}
\end{figure}

\begin{figure}[t]
    \centering
    \includegraphics[width=0.8\textwidth]{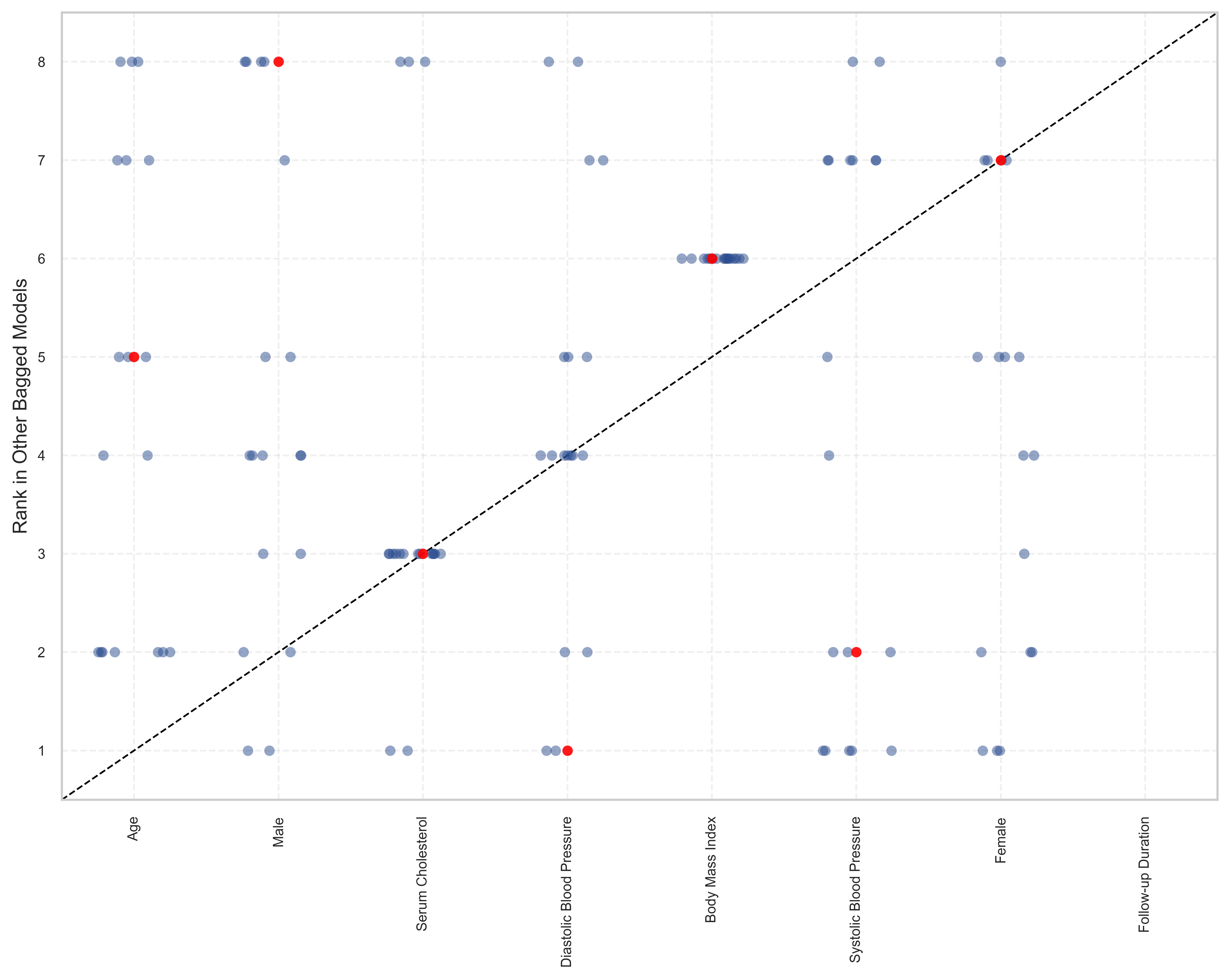}
    \caption{{\bf Feature attribution rankings.} Variability in SHAP-based feature attribution rankings across the constituents of the \ensemble model for the Framingham dataset. {\color{red}Red} dots indicate the rankings of one randomly chosen constituent model for reference.}
    \label{fig:framranking}
\end{figure}

In the experiments, bagging models achieved maximal ILP stability at the expense of model interpretation which is demonstrate below. Figure \ref{fig:ensembleshap} shows the variation in SHAP values across ensemble model constituents for each of the real data sets. The long tails in the SHAP distribution are due to constituent models placing importance on different predictors. Looking in particular at the Framingham data set, Figure \ref{fig:framranking} shows the variation of the ranking of SHAP value importance for each variable. While in one model, diastolic and systolic blood pressure were ranked as the two most important predictors, in other models, the predictors had lower importance and were bottom ranked in some models.

\end{document}